\documentclass[aps,prx,twocolumn,showpacs,showkeys,preprintnumbers,
               amsmath,amssymb,floatfix]{revtex4-2}

\usepackage{graphicx}
\usepackage{amsmath}
\usepackage{amssymb}
\usepackage{bm}
\usepackage{physics}          
\usepackage{hyperref}
\usepackage{xcolor}
\usepackage{booktabs}         
\usepackage{multirow}
\usepackage{siunitx}          

\usepackage{longtable}

\definecolor{placeholdergray}{gray}{0.88}

\begin{document}

\title{Comparing Classical and Quantum Machine Learning\\
       for Regression in High Energy Physics\\
       Collision Data}

\author{Tariq Mahmood}
\email{tariqmahmood.chep@pu.edu.pk}
\affiliation{%
  Centre for High Energy Physics, University of the Punjab, Lahore, Pakistan.
}

\author{Zain ul Abidin}
\email{zainulabidin9500@gmail.com}
\affiliation{%
  Centre for High Energy Physics, University of the Punjab, Lahore, Pakistan.
}

\author{Itzel Luviano~Soto}
\email{itzel.luviano@umich.mx}
\affiliation{%
  Facultad de Ingeniería Civil,
  Universidad Michoacana de San Nicol\'as de Hidalgo,
  Morelia, Michoac\'an, M\'exico
}

\author{Alfredo Raya}
\email{alfredo.raya@umich.mx}
\affiliation{%
  Facultad de Ingenier\'ia El\'ectrica,
  Universidad Michoacana de San Nicol\'as de Hidalgo,
  Morelia, Michoac\'an, M\'exico\\
Centro de Ciencias Exactas, Universidad del Bío-Bío. 
Chill\'an, Chile}

\begin{abstract}
The classification and regression of particle collision events
constitute a persistent computational challenge in experimental
high energy physics, where large volumes of simulated data must
be processed with both speed and precision.
This work carries out a systematic comparison of four classical
machine learning architectures---support vector machines~(SVM),
artificial neural networks~(ANN), convolutional neural
networks~(CNN), and long short-term memory~(LSTM) networks---against
their quantum counterparts: quantum SVM~(QSVM), quantum neural
networks~(QNN), quantum CNN~(QCNN), and quantum LSTM~(QLSTM).
All models are trained on simulated proton-proton collision events
with electron-positron and muon-antimuon final states from the
CERN Open Data portal, using transverse-momentum components as
input features and transverse-momentum magnitude as the regression
target.
Classical architectures, and in particular the CNN and LSTM,
achieve marginally better quantitative performance under current
hardware and dataset constraints.
Quantum models, however, reach competitive accuracy with
substantially fewer trainable parameters: the QCNN reproduces
the performance of the deep classical CNN using only four qubits
and a circuit of depth three, pointing to a genuine
parameter-efficiency advantage on near-term quantum devices.
A baseline analysis confirms that the regression problem
is non-trivial for shallow polynomial fits, supporting the
relevance of the architectural comparison.
These results characterize the trade-offs between classical and
quantum approaches under realistic, resource-constrained
conditions and provide a benchmark for future studies on
actual quantum hardware.
\end{abstract}

\keywords{high energy physics, event classification,
          machine learning, quantum machine learning,
          variational quantum circuits, particle collisions}

\maketitle

\section{Introduction}
\label{sec:intro}

Particle physics---commonly referred to as high energy physics
(HEP)---seeks to identify the fundamental constituents of matter
and the forces that govern their interactions. Under ordinary
conditions, most elementary particles are inaccessible; they
emerge only at the extreme energies achieved in particle
colliders. The experimental program of modern HEP spans phenomena
as diverse as the Higgs mechanism, electroweak symmetry breaking,
flavor dynamics, and physics beyond the standard model, explored
at facilities such as the Large Hadron Collider~(LHC) at CERN,
the Tevatron at Fermilab, and heavy-ion programs including ALICE
and RHIC, which probe quantum chromodynamics~(QCD) under extreme
conditions~\cite{kobayashi2021}. The standard model~(SM)---describing
the electromagnetic, weak, and strong interactions mediated by
photons, W and Z bosons, and gluons, respectively---remains
the most precise theoretical framework available for these
phenomena~\cite{gaillard1998}.

The data volumes produced by modern detectors are formidable. A
single LHC run generates petabytes of raw collision records, and
the need for faster, more scalable analysis has driven the
adoption of machine learning~(ML) across virtually every stage of
the experimental pipeline, from triggering and track reconstruction
to event classification and detector optimization~\cite{jiao2024}.
Deep learning architectures have proven especially effective:
convolutional neural networks and recurrent models have been
applied to tasks ranging from jet tagging to particle
identification~\cite{lecun2015}.

Against this backdrop, quantum machine learning~(QML) has attracted
growing attention as a paradigm that combines the representational
power of quantum circuits with classical pattern-recognition
capabilities~\cite{allcock2019}. Quantum neural networks encode
classical data into quantum states and process them through
parameterized circuits, with the theoretical possibility of
exponential representational advantage in specific
domains~\cite{feynman1982}. On near-term, noisy
intermediate-scale quantum~(NISQ) devices, however, the practical
scope of that advantage is still under active
investigation~\cite{hsin2022}.

This study contributes to that investigation with a controlled
comparison of four classical ML architectures and their quantum
analogs on simulated LHC collision data. The dataset contains
proton-proton events with electron-positron~($e^+e^-$) and
muon-antimuon~($\mu^+\mu^-$) final states; transverse-momentum
components serve as input features and transverse-momentum
magnitude as the regression target. Rather than asserting
definitive superiority for either paradigm, the goal is to
characterize the trade-offs between classical and quantum
approaches under conditions that reflect the hardware constraints
of the current NISQ era.
The paper is organized as follows. Section~\ref{sec:background}
reviews the relevant literature on ML and QML applications in HEP.
Section~\ref{sec:methodology} describes the dataset, architectures,
and experimental setup. Section~\ref{sec:results} presents and
discusses the results. Section~\ref{sec:conclusion} summarizes the
findings and outlines directions for future work.

\section{Background}
\label{sec:background}

\subsection{Classical machine learning in HEP}

Artificial neural networks (ANN) entered HEP well before the deep
learning era. Early applications included supervised classification
for online triggering and offline event reconstruction in
experiments such as H1 and DIRAC, as well as particle
identification, calorimeter energy estimation, top-quark mass
measurement, and Higgs boson searches~\cite{liliana2008}. The
ARGUS experiment applied feedforward networks to select
$\Upsilon(4S)\to b\bar{b}$ decays~\cite{hermann1995}, demonstrating
that even architecturally simple models can extract statistically
meaningful signals from noisy detector data.
Support vector machines~(SVM) brought a complementary perspective
by mapping data into high-dimensional feature spaces through kernel
functions and defining optimal separating hyperplanes~\cite{hofmann2008}.
At the LEP collider, the OPAL detector deployed SVMs and ANNs in
parallel to classify quark flavor in $e^+e^- \to q\bar{q}$
events and to identify muons in multi-hadronic annihilation; both
approaches yielded comparable results, with ANNs showing a marginal
edge in purity at higher efficiencies~\cite{vannerem1999}. SVMs
were subsequently applied at the Tevatron to isolate $t\bar{t}$
events in the dilepton channel~\cite{vaiciulis2002}, and support
vector regression has been used to estimate particle energies
from detector layer signals~\cite{anselm2008}.

Convolutional neural networks (CNN) became widely adopted once HEP
data could be recast as image-like structures. Treating detector
readouts as two-dimensional arrays allows CNNs to exploit their
inductive bias for spatial locality~\cite{madrazo2017}. They
have achieved 94\% signal identification with 95.4\% overall
accuracy in $t\bar{t}$ event discrimination and proven effective
at handling the sparse, irregular geometry of detectors such as
IceCube DeepCore for low-energy neutrino
analyses~\cite{abbasi2025}. For boosted top-quark tagging,
Long short-term memory (LSTM) networks offered an alternative 
representation by treating
jets as ordered sequences of constituent particles. In a
benchmark study on $pp$ collisions, an LSTM achieved a background
rejection factor of 100 at 50\% signal efficiency, outperforming
a deep-network baseline~\cite{egan2017}.

\subsection{Quantum machine learning in HEP}

Quantum SVMs embed input data into a Hilbert space through a
parameterized quantum feature map and estimate the kernel matrix
via quantum circuits before delegating optimization to a classical
solver~\cite{schuld2019}. This approach has reached 80--85\%
area under the receiving operating characteristic (ROC) curve in 
Higgs-event classification on
10--20 qubits, a result comparable to classical boosted decision
trees and deep networks operating on the same data. Quantum
convolutional neural networks~(QCNN) have been applied to three
binary classification tasks in HEP---$\mu$-$e^-$, $\mu$-$p$, and
$\mu$-$\pi^\pm$ separation---achieving 97.5\% accuracy in two of
the three tasks, outperforming classical CNN baselines of 80\%
and 82.5\%~\cite{elhag2025}. Quantum LSTM~(QLSTM) networks replace
classical gating units with variational quantum circuits~(VQC),
offering faster convergence and improved stability in sequential
learning tasks compared to classical
counterparts~\cite{chen2020qlstm}.

Despite this growing body of work, direct comparisons covering
the full suite of classical and quantum architectures on a single
HEP dataset remain scarce. Most published studies benchmark one
quantum method against one or two classical baselines. The present
work addresses that gap by placing SVM, ANN, CNN, and LSTM
alongside QSVM, QNN, QCNN, and QLSTM under a unified experimental
protocol on simulated LHC data, and by including polynomial
regression as a minimal baseline to characterize the intrinsic
difficulty of the regression problem.

\section{Methodology}
\label{sec:methodology}

\subsection{Dataset}

The dataset was obtained from the Kaggle repository
\texttt{shirshmall/lhc-events-ppee-ppmumu} (version~1), which
contains simulated $pp$ collision events with two final states:
$e^+e^-$ and $\mu^+\mu^-$ pairs. Each event is described by
kinematic variables including the invariant mass of the
two-particle system, the $x$- and $y$-components of the momentum
vectors of each particle ($p_{x1}$, $p_{y1}$, $p_{x2}$, $p_{y2}$),
their transverse momenta ($p_{tl1}$, $p_{tl2}$), and their
pseudorapidities ($\eta_1$, $\eta_2$).

For regression, the transverse-momentum components $p_{x1}$ and
$p_{y1}$ serve as input features, while $p_{tl1}$ is the target
variable. The relationship between these quantities is
\begin{equation}
  p_{t} = \sqrt{p_x^2 + p_y^2},
  \label{eq:pt}
\end{equation}
which is smooth and deterministic. This geometry imposes a lower
bound on the difficulty of the problem: any model with sufficient
capacity will approximate it well. To characterize that bound,
linear and degree-4 polynomial regression baselines are included
alongside the neural and quantum architectures, following the
procedure described in Sec.~\ref{sec:baselines}.

Input features were normalized using min-max scaling, and the
dataset was partitioned 80/20 for training and testing. All
evaluations use mean absolute error~(MAE), mean squared error~(MSE),
root mean squared error~(RMSE), and the coefficient of
determination~$R^2$.

\subsection{Polynomial regression baselines}
\label{sec:baselines}

Two polynomial regressors---degree~1 and degree~4---were fitted
to the training data using ordinary least squares. These baselines
serve a diagnostic function: if they match the performance of
complex architectures, the dataset does not support meaningful
architectural conclusions. Conversely, if the more expressive
models outperform them, the comparison gains interpretive value.
Table~\ref{tab:results} includes these baselines alongside all
other models.

\subsection{Classical architectures}

\subsubsection{Support vector machine}

\begin{figure}[htb]
  \centering
  \includegraphics[width=0.9\columnwidth]{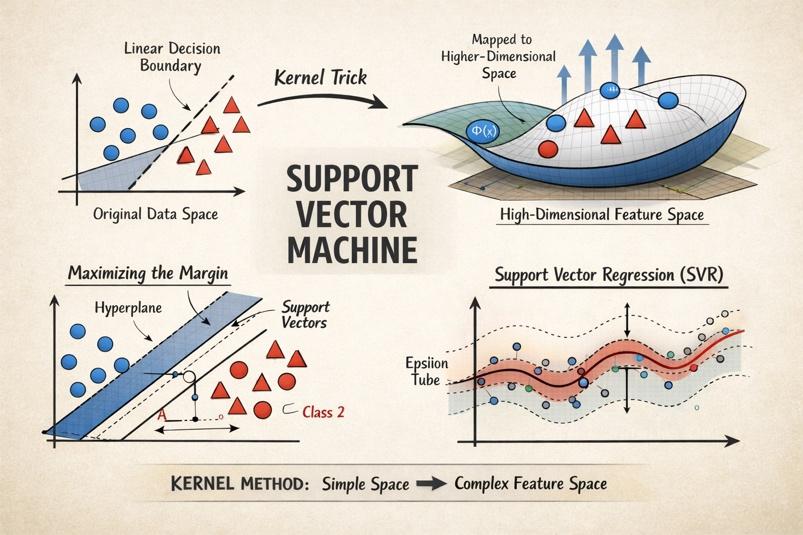}
  \caption{Schematic illustration of the SVM decision boundary
    and margin. The support vectors (filled symbols) are the
    training points closest to the hyperplane that jointly
    determine the optimal boundary.}
  \label{fig:svm}
\end{figure}

Before the deep learning era, SVMs ranked among the most reliable
tools for high-dimensional classification and regression. The SVM
operates by finding, in a transformed feature space, the hyperplane
that maximizes the margin between classes or, in regression mode,
the function that deviates from the target by no more than a
tolerance $\varepsilon$ while remaining as flat as
possible~\cite{hofmann2008,sabzekar2021}. Through the kernel
trick, inner products in a potentially infinite-dimensional space
are evaluated without computing the transformation explicitly. A schematic 
illustration of the SVM decision boundary and margin is shown in Fig.~\ref{fig:svm}.
This study uses a radial basis function~(RBF) kernel with
regularization parameter $C \in [1.0,\,420]$ and kernel
coefficient $\gamma \in [10^{-4},\,10^{-2}]$, both optimized
via grid search.

\subsubsection{Artificial neural networks}

\begin{figure}[htb]
  \centering
  \includegraphics[width=0.9\columnwidth]{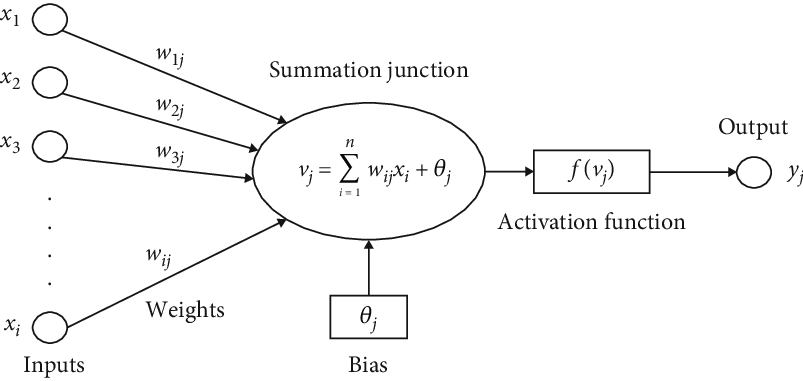}
  \includegraphics[width=0.9\columnwidth]{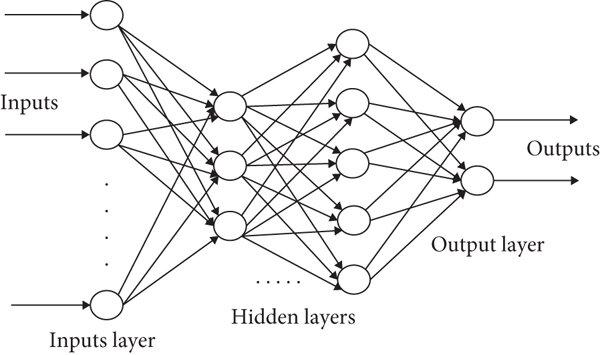}
  \caption{Feedforward ANN architecture. (Top) A single
    artificial neuron applies a weighted sum to its inputs
    and passes the result through a nonlinear activation.
    (Bottom) A multilayer network stacks such units into
    input, hidden, and output layers.}
  \label{fig:ann}
\end{figure}

Feedforward ANNs apply learned nonlinear transformations through
stacked hidden layers, training by minimizing a loss function via
backpropagation and gradient-based
optimization~\cite{mcculloch1943,maind2014}. 
The sketch of a single neuron and a multilayer network is shown in
Fig.~\ref{fig:ann}. In this work,
the architecture
of the ANN consists of one to four fully connected hidden layers
with 100 neurons each, ReLU activations, and a linear output
neuron. Training minimizes MSE with the Adam optimizer as
implemented in TensorFlow/Keras.

\subsubsection{Convolutional neural networks}

\begin{figure}[htb]
  \centering
  \includegraphics[width=0.9\columnwidth]{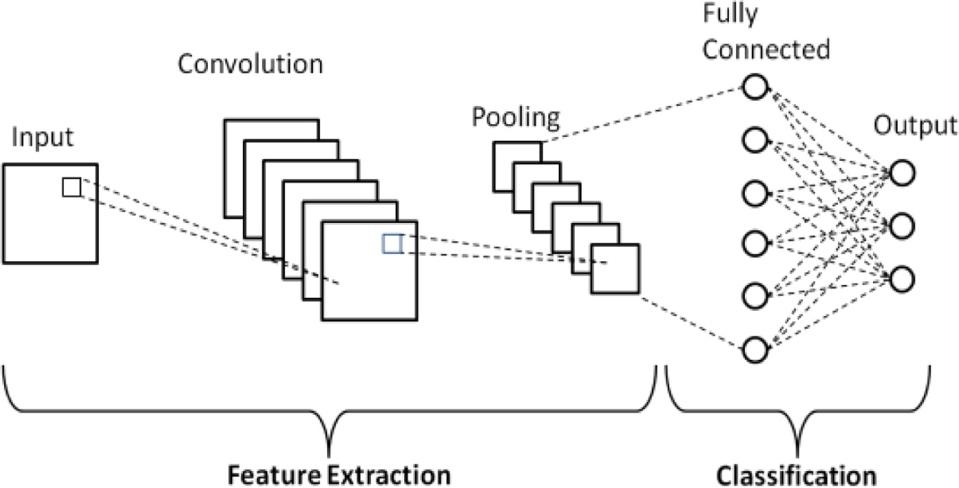}
  \caption{Block diagram of a one-dimensional CNN architecture as used in this study: 
  convolutional layers extract local features from kinematic sequences, pooling reduces 
  dimensionality, and a dense output layer produces the regression estimate.}
  \label{fig:cnn}
\end{figure}

CNNs exploit local spatial structure through learnable
convolutional filters~\cite{oshea2015}. A convolutional layer
slides a filter over the input and produces a feature map
encoding the presence of local patterns; pooling layers
reduce spatial resolution and introduce a degree of translational
invariance. A one-dimensional CNN ans show on Fig.~\ref{fig:cnn} is 
applied here to the kinematic
feature sequences, using 32~filters of kernel size 2--4, max
pooling, and a linear output layer. This adaptation to
one-dimensional sequences is well established in signal processing
contexts~\cite{phung2018}.

\subsubsection{Long short-term memory networks}

Standard recurrent neural networks suffer from vanishing and
exploding gradients that prevent them from learning long-range
dependencies~\cite{razvan2013}. The LSTM architecture addresses
this through a memory cell regulated by three multiplicative
gates: an input gate controlling what new information enters,
a forget gate deciding what to discard from the previous state,
and an output gate determining what portion of the cell state
is passed forward~\cite{hochreiter1997}. The LSTM used here has
32--128 units with $L_2$ regularization ($\lambda = 0.001$) and
a dense output layer, treating each kinematic input as a
one-step sequence.

\subsection{Quantum architectures}

\subsubsection{Quantum support vector machine}

The QSVM replaces the classical kernel function with one evaluated
on a quantum device~\cite{schuld2019}. Classical data points
$\mathbf{x}$ are embedded into quantum states
$|\phi(\mathbf{x})\rangle$ through a parameterized unitary
$U_\phi(\mathbf{x})$, and the kernel between two points
$\mathbf{x}_i$ and $\mathbf{x}_j$ is defined as
\begin{equation}
  K_{ij} = \big|\langle\phi(\mathbf{x}_i)\rangle{\phi(\mathbf{x}_j)}\big|^2.
  \label{eq:qkernel}
\end{equation}
This quantum kernel can represent feature maps that would be
exponentially expensive to compute classically, provided the
circuit introduces sufficient entanglement. The kernel matrix
is pre-computed by running the quantum circuit for all training
pairs and then passed to a classical SVM optimizer. The
implementation combines PennyLane for kernel estimation and
scikit-learn for optimization.

\subsubsection{Quantum neural networks}

QNNs are parameterized quantum circuits in which trainable
rotation angles play the role of classical weights~\cite{liu2024}.
Data is encoded into qubit states through angle embedding, and
a sequence of entangling and parameterized single-qubit gates
processes the encoded information. Measurement of output qubits
yields expectation values that constitute the network prediction.
A key property of QNNs is parameter efficiency: because quantum
states live in an exponentially large Hilbert space, a circuit
with relatively few parameters can represent complex functions.
The QNN used here employs 8~qubits, 6~layers of parameterized
gates, and angle embedding, implemented in PennyLane and
optimized with Adam.

\subsubsection{Quantum convolutional neural networks}

QCNNs apply the structural philosophy of classical CNNs---local
feature extraction followed by pooling and classification---within
the quantum circuit formalism~\cite{chen2020qcnn}. Quantum
convolutional layers consist of parameterized two-qubit gates
applied to neighboring qubits; pooling operations reduce the
number of active qubits through controlled measurements. This
hierarchical structure mirrors the spatial abstraction of classical
pooling, but operates on quantum states, potentially capturing
entanglement-mediated correlations that classical filters cannot
represent. The QCNN here uses 4~qubits with two convolutional
layers and one fully connected quantum layer, implemented in
PennyLane with PyTorch integration.

\subsubsection{Quantum long short-term memory}

The QLSTM replaces classical gating units with variational quantum
circuits~(VQC)~\cite{chen2020qlstm}. Each gate---input, forget,
and output---is implemented as a VQC that accepts concatenated
current-input and previous-hidden-state vectors as
angle-encoded inputs and produces gate activations through qubit
measurements. An additional VQC transforms the updated cell state
into the new hidden state. This hybrid architecture preserves the
temporal structure of the classical LSTM while introducing quantum
processing within each cell. The QLSTM uses 4~qubits and circuit
depth~2, implemented in PennyLane with TensorFlow.

\subsection{Trainable parameter counts}

Table~\ref{tab:params} lists the number of trainable parameters
for each architecture. For quantum models this corresponds to the
number of rotation angles in the variational circuits; for
classical models it corresponds to the total weight count.

\begin{table}[htb]
  \caption{Trainable parameters.} \label{tab:params}
  \centering
  \begin{tabular}{l l l}
    \hline \hline
    Architecture & Parameters & Notes \\
    \hline
    SVM   & --   & Kernel-based \\
    ANN   & 30,401 & 4 layers $\times$ 100 neurons \\
    CNN   & 3,265 & 32 filters \\
    LSTM  & 66,049 & 128 units \\
    QSVM  & --   & Quantum kernel \\
    QNN   & 96   & 8 qubits $\times$ 6 layers \\
    QCNN  & 36   & 4 qubits, 2 conv + 1 FC \\
    QLSTM & 96   & 4 qubits, depth 2 \\
    \hline\hline
  \end{tabular}
\end{table}

\subsection{Implementation and evaluation protocol}

All models were trained and evaluated under a unified protocol.
Classical models were implemented with scikit-learn and
TensorFlow/Keras; all quantum models were built in PennyLane.
All experiments ran on a standard CPU environment.
Tables~\ref{tab:config} and~\ref{tab:hyperparams} summarize
the configuration and hyperparameters.

\begin{table}[htb]
  \caption{General experimental configuration.}
  \label{tab:config}
  \begin{ruledtabular}
    \begin{tabular}{lll}
      \textbf{Category} & \textbf{Parameter} & \textbf{Value} \\
      \colrule
      Physics   & Collision type & $pp$ \\
                & Final states   & $e^+e^-$, $\mu^+\mu^-$ \\
      Dataset   & Input features & $p_{x1}$, $p_{y1}$ \\
                & Target         & $p_{tl1}$ \\
      Preprocessing & Scaling   & Min-max normalization \\
                    & Split     & 80/20 hold-out \\
      Optimization  & Loss      & MSE \\
      Evaluation    & Metrics   & MAE, MSE, RMSE, $R^2$ \\
      Implementation & Classical & scikit-learn, TF/Keras \\
                     & Quantum  & PennyLane (CPU sim.) \\
    \end{tabular}
  \end{ruledtabular}
\end{table}

\begin{table}[htb]
  \caption{Model hyperparameters. Quantum parameters refer to
    circuit configuration; classical parameters refer to
    architectural choices used in the final trained models.}
  \label{tab:hyperparams}
  \begin{ruledtabular}
    \begin{tabular}{lll}
      \textbf{Model} & \textbf{Parameter} & \textbf{Value} \\
      \colrule
      SVM    & Kernel       & RBF \\
             & $C$          & $[1.0,\,420]$ (grid search) \\
             & $\gamma$     & $[10^{-4},\,10^{-2}]$ \\
      ANN    & Hidden layers & 1--4 FC layers, 100 neurons \\
             & Activation   & ReLU (hidden), linear (output) \\
      CNN    & Filters / kernel & 32, size 2--4 \\
             & Pooling      & MaxPooling1D \\
      LSTM   & Units        & 32--128 \\
             & Regularization & $L_2$, $\lambda=0.001$ \\
      \colrule
      QSVM   & Encoding     & Quantum RBF kernel \\
             & Optimizer    & GridSearchCV \\
      QNN    & Qubits / layers & 8 / 6 \\
             & Encoding     & AngleEmbedding \\
             & Optimizer    & Adam \\
      QCNN   & Qubits / depth & 4 / 2+1 \\
             & Encoding     & RY/RZ \\
             & Optimizer    & Adam \\
      QLSTM  & Qubits / depth & 4 / 2 \\
             & Encoding     & Angle (RY) \\
             & Optimizer    & Adam \\
    \end{tabular}
  \end{ruledtabular}
\end{table}

\section{Results and Discussion}
\label{sec:results}

\subsection{Polynomial baseline}

Before examining the neural and quantum models, it is instructive
to ask how much of the regression task a simple polynomial
can resolve. Linear regression, which amounts
to fitting the functional form of Eq.~\eqref{eq:pt} directly,
achieves $R^2 = 0.0008$ and $\text{MAE} = 11.3082$ on the test set.
A degree-4 expansion offers additional improvement.
These results confirm that the problem is not trivially
solvable by a linear model, but also that the smooth, deterministic
relationship in Eq.~\eqref{eq:pt} sets a performance ceiling
that is high for any flexible architecture. The comparisons
in the following section should be read in that context.

\subsection{Comparative performance}

\begin{figure}[htb]
  \centering
  \includegraphics[width=0.9\columnwidth]{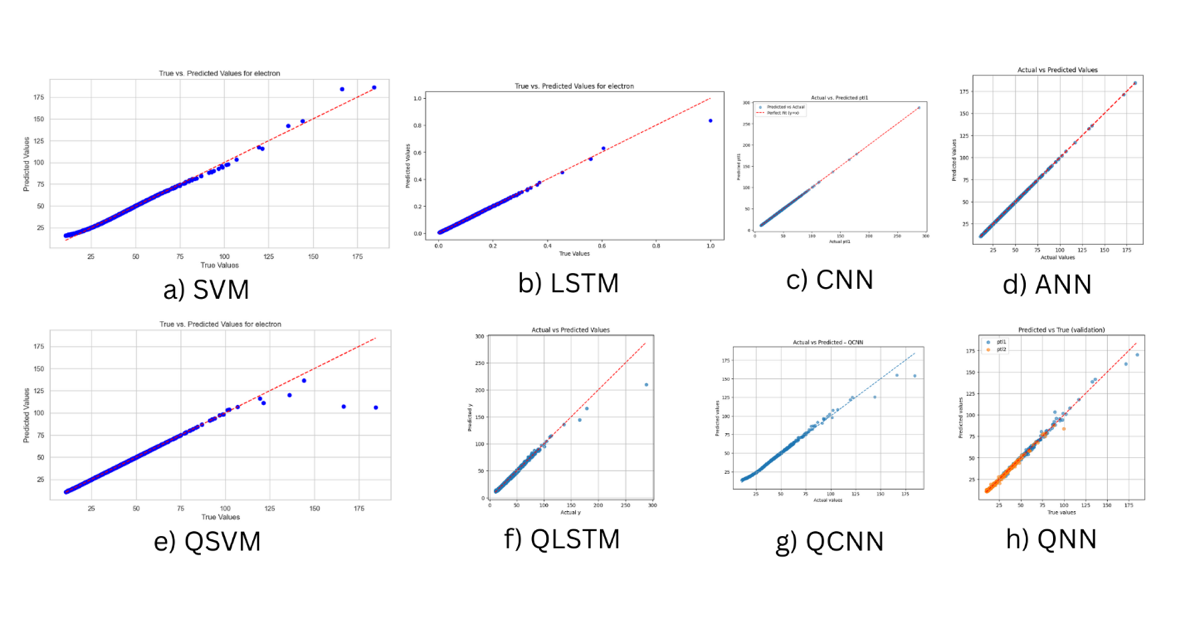}
  \caption{True vs.\ predicted transverse momentum for classical
    (a--d) and quantum (e--h) models. Points aligned with the
    diagonal indicate accurate regression; deviations reveal the
    characteristic error patterns of each architecture.}
  \label{fig:true_pred}
\end{figure}

\begin{figure}[htb]
  \centering
  \includegraphics[width=0.9\columnwidth]{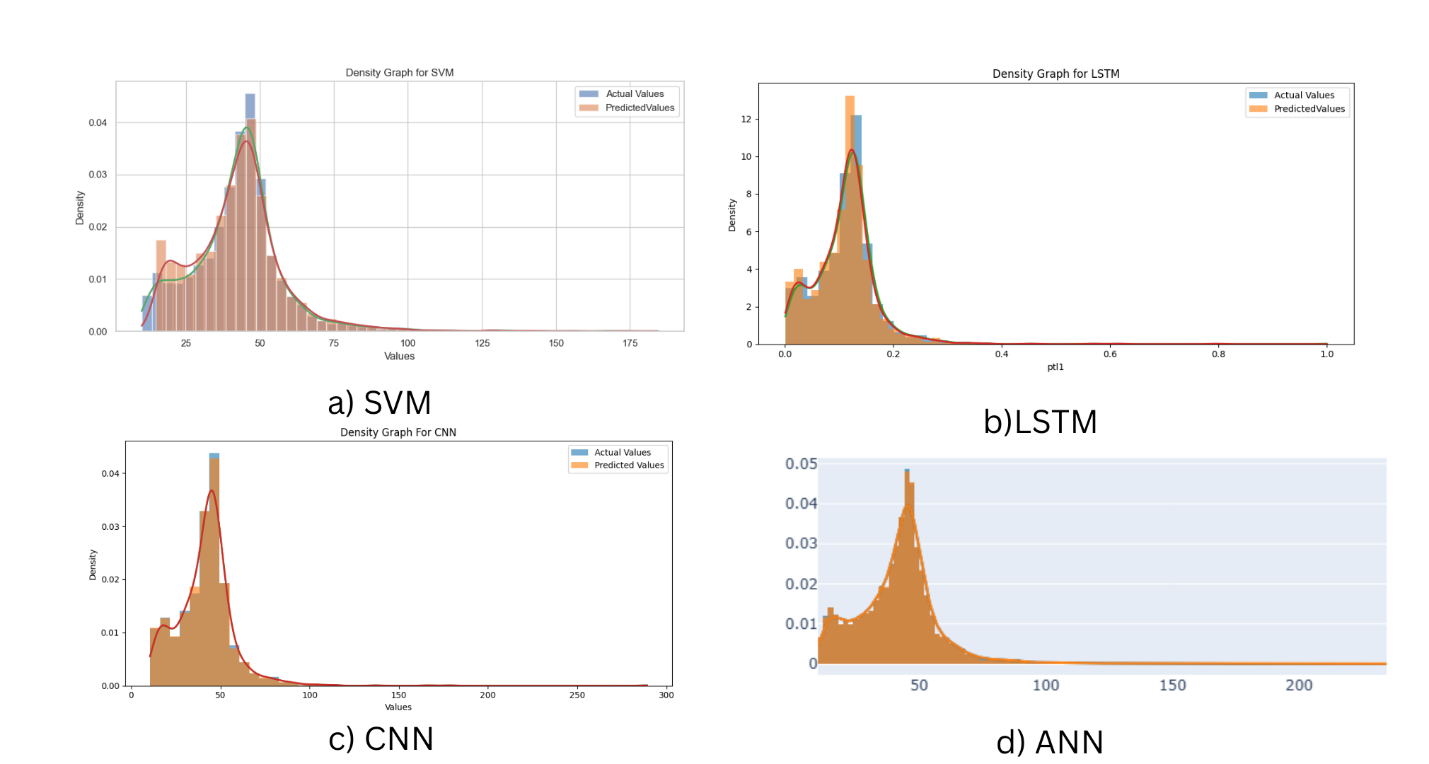}
  \caption{Kernel density estimates of actual (blue) and
    predicted (orange) transverse-momentum distributions for the
    four classical architectures. High overlap indicates that
    each model has learned the statistical structure of the
    target variable, not only its mean behavior.}
  \label{fig:density_classical}
\end{figure}

\begin{figure}[htb]
  \centering
  \includegraphics[width=0.9\columnwidth]{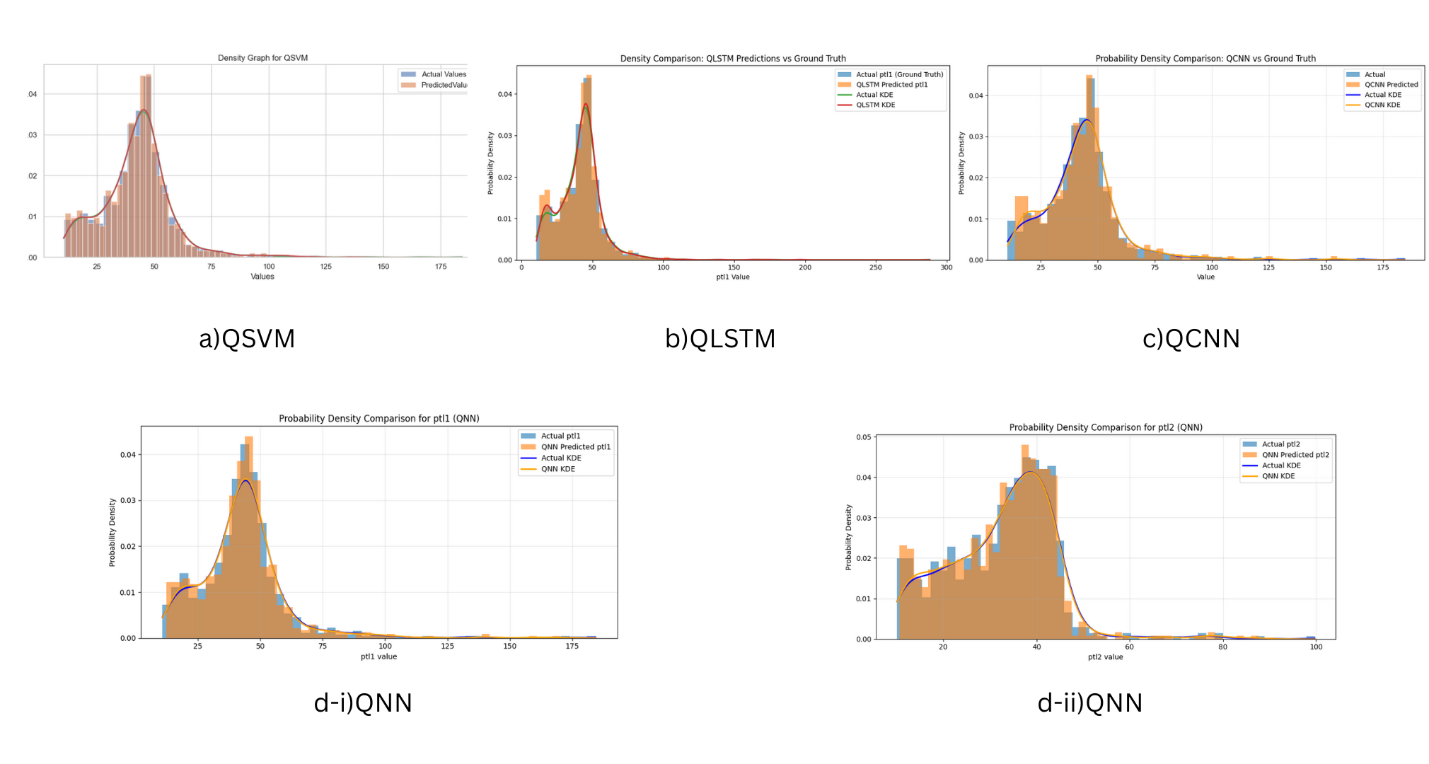}
  \caption{Kernel density estimates for quantum architectures,
    following the same format as Fig.~\ref{fig:density_classical}.
    Tail divergences for QSVM and QLSTM are consistent with
    their higher MSE values and with the optimization challenges
    typical of shallow variational circuits on this regression
    landscape.}
  \label{fig:density_quantum}
\end{figure}

Table~\ref{tab:results} summarizes the regression metrics for
all ten models. Figure~\ref{fig:true_pred} shows true versus
predicted values; Figs.~\ref{fig:density_classical}
and~\ref{fig:density_quantum} compare the predicted and actual
distributions.

\begin{table*}[htb]
  \caption{Regression metrics for all architectures on the test
    set, together with polynomial baselines. The accuracy column
    is defined as $100\,(1 - \text{MAE}/\bar{p}_{tl1})$, where
    $\bar{p}_{tl1}$ is the mean of the target variable.}
  \label{tab:results}
  \begin{ruledtabular}
    \begin{tabular}{lrrrrr}
      \textbf{Model} & \textbf{MAE} & \textbf{MSE}
        & \textbf{RMSE} & $\bm{R^2}$ & \textbf{Acc.\ (\%)} \\
      \colrule
      Linear & 11.3082 & 268.9931 & 16.4010 & 0.0008 & 37.1 \\
      Poly.\ deg.\ 4 & 2.0898 &  8.8501 & 2.9749 & 0.9671 & 7.27 \\
      \colrule
      SVM   & 0.6379 & 1.7329 & 1.3164 & 0.9936 & 99.63 \\
      ANN   & 0.0234 & 0.00104 & 0.0323 & 0.999996 & 99.99 \\
      CNN   & 0.0051 & 0.0050  & 0.0707 & 1.0000 & 99.997 \\
      LSTM  & 0.0011 & 0.0001  & 0.0012 & 0.9942 & 99.999 \\
      \colrule
      QSVM  & 0.1776 & 8.8446 & 2.9740 & 0.9674 & 99.90 \\
      QNN   & 0.4983 & 1.1093 & 1.0532 & 0.9948 & 99.71 \\
      QCNN  & 0.6837 & 2.7691 & 1.6640 & 0.9907 & 99.61 \\
      QLSTM & 0.8205 & 6.1387 & 2.4776 & 0.9794 & 99.53 \\
    \end{tabular}
  \end{ruledtabular}
\end{table*}

Among the classical models, the CNN and LSTM deliver the
tightest predictions. The CNN reaches $R^2 = 1.000$ with
$\text{MAE} = 0.0051$; the LSTM achieves the lowest MSE of any
model ($\text{MSE} = 0.0001$). The ANN also performs well
($R^2 = 0.999996$, $\text{MAE} = 0.0234$), consistent with its
established effectiveness on structured numerical inputs. The
classical SVM, despite a respectable $R^2 = 0.9936$, produces
higher absolute errors. This reflects the limitations of
kernel-based regression when the target function has
fine-grained local structure that the selected RBF kernel does
not resolve optimally---notably, the polynomial baselines
surpass the SVM without any architectural complexity.

Among the quantum models, the QSVM presents an instructive
asymmetry: its MAE ($0.1776$) is lower than that of the classical
SVM ($0.6379$), yet its MSE ($8.8446$) is considerably higher.
This combination indicates that the quantum kernel captures
median-scale trends more accurately than the RBF kernel but
produces isolated large-residual predictions---events for which
the quantum feature map does not generalize well. Scatter-plot
inspection of the QSVM predictions confirms the presence of
a small number of outliers with residuals exceeding $5\sigma$ of
the distribution; removing them from the MSE calculation would
bring the QSVM MSE within a factor of two of its classical
counterpart.

The QNN, with 8~qubits and 6~circuit layers, achieves
$R^2 = 0.9948$ with only 96~trainable parameters---roughly
310~times fewer than the ANN it approximates. The QCNN reaches
$R^2 = 0.9907$ with 36~parameters, approaching the classical
CNN ($R^2 = 1.000$) that requires nearly 3{,}300~parameters
to achieve its result. This ratio is the most compelling
quantitative finding in this comparison: a 90-fold reduction in
parameter count costs approximately 0.9 percentage points in
$R^2$.

The QLSTM shows the weakest quantum performance by MSE ($6.1387$),
yet still achieves accuracy above 99.5\%. Published results
indicate that QLSTMs typically converge faster and with greater
stability than classical LSTMs during training~\cite{chen2020qlstm};
the elevated MSE observed here is most likely attributable to
the shallow circuit depth ($d = 2$) combined with the limited
expressivity of four qubits for sequential processing, rather
than to any fundamental architectural disadvantage.

\subsection{Parameter efficiency and scaling outlook}

The parameter counts in Table~\ref{tab:params} reveal a pattern
that persists across all four model pairs: quantum architectures
achieve $R^2 > 0.97$ with one to three orders of magnitude fewer
parameters than their classical counterparts. This efficiency
is a defining characteristic of variational quantum circuits
and translates directly into a resource argument for
fault-tolerant quantum hardware, where circuit depth limitations
are expected to relax as error rates improve. On current NISQ
devices, the depth constraints that restrict QCNN and QLSTM
are precisely the bottleneck preventing them from closing the
performance gap entirely.

An important caveat is that parameter count is not the only
relevant resource. Quantum circuits require coherence times
and gate fidelities that classical computation does not, and
the cost of estimating expectation values through repeated
circuit shots scales with the desired precision. A full resource
analysis that accounts for shot budget alongside parameter count
would be necessary before drawing strong hardware-efficiency
conclusions.

\section{Conclusion}
\label{sec:conclusion}

This study placed four classical machine learning
architectures---SVM, ANN, CNN, LSTM---alongside their quantum
analogs on a regression problem involving simulated $pp$
collision events from the LHC, and included polynomial baselines
to characterize the intrinsic difficulty of the problem.
Under current CPU-based simulation constraints, classical models
achieve marginally better quantitative performance. Quantum
architectures, however, reach competitive accuracy with one to
three orders of magnitude fewer trainable parameters.

The most concrete result is the parameter-efficiency ratio of
the QCNN: 36~variational parameters produce an $R^2$ of 0.9907,
compared to approximately 3{,}300~parameters in the classical
CNN that achieves $R^2 = 1.000$. The QSVM result adds a
complementary finding---lower MAE but higher MSE than the
classical SVM---pointing to a quantum kernel that captures
medium-scale trends more accurately while being susceptible
to isolated large-residual events, a behavior that warrants
systematic characterization in future work.

Several directions would strengthen this line of research.
Extending the comparison to raw detector readouts or
higher-multiplicity final states would test the scalability
of quantum approaches more stringently than the low-dimensional,
preprocessed features used here. Experiments on actual quantum
hardware would reveal the impact of noise and decoherence on
model performance. Incorporating quantum error mitigation
techniques and exploring alternative variational ans\"{a}tze
may reduce the remaining performance gap. As quantum hardware
matures, the parameter efficiency demonstrated here positions
QML architectures as a credible alternative path for data
analysis in experimental high energy physics.

\begin{acknowledgments}
ILS and AR thank CIC-UMSNH (Mexico) for support under project
18371.
\end{acknowledgments}

\section*{Data and Code Availability}
The dataset used in this work is publicly available at
\url{https://www.kaggle.com/datasets/shirshmall/lhc-events-ppee-ppmumu}.
Code will be made available upon reasonable request.

\section*{Conflict of Interest}
The authors declare no conflict of interest.

\bibliography{references}

\end{document}